\documentclass[conference]{IEEEtran}
\IEEEoverridecommandlockouts
\usepackage{cite}
\usepackage{amsmath,amssymb,amsfonts}
\usepackage{algorithmic}
\usepackage{graphicx}
\usepackage{textcomp}
\usepackage{xcolor}
\usepackage{cite}
\usepackage{url}
\def\BibTeX{{\rm B\kern-.05em{\sc i\kern-.025em b}\kern-.08em
    T\kern-.1667em\lower.7ex\hbox{E}\kern-.125emX}}
\begin{document}

\title{LiteLSTM Architecture for Deep Recurrent Neural Networks%\\
%{\footnotesize \textsuperscript{*}Note: Sub-titles are not captured in Xplore and
%should not be used}
%\thanks{Identify applicable funding agency here. If none, delete this.}
}

\author{\IEEEauthorblockN{Nelly Elsayed$^{\dagger}$, Zag ElSayed$^{\dagger}$, Anthony S. Maida$^{\mathsection}$}
	\IEEEauthorblockA{\textit{$^{\dagger}$School of Information Technology} \\
		$^{\mathsection}$\textit{School of Computing and Informatics}\\
		$^{\dagger}$University of Cincinnati\\
		$^{\mathsection}$University of Louisiana at Lafayette}
	}
%\author{\IEEEauthorblockN{1\textsuperscript{st} Given Name Surname}
%\IEEEauthorblockA{\textit{dept. name of organization (of Aff.)} \\
%\textit{name of organization (of Aff.)}\\
%City, Country \\
%email address or ORCID}
%\and
%\IEEEauthorblockN{2\textsuperscript{nd} Given Name Surname}
%\IEEEauthorblockA{\textit{dept. name of organization (of Aff.)} \\
%\textit{name of organization (of Aff.)}\\
%City, Country \\
%email address or ORCID}
%\and
%\IEEEauthorblockN{3\textsuperscript{rd} Given Name Surname}
%\IEEEauthorblockA{\textit{dept. name of organization (of Aff.)} \\
%\textit{name of organization (of Aff.)}\\
%City, Country \\
%email address or ORCID}
%\and
%\IEEEauthorblockN{4\textsuperscript{th} Given Name Surname}
%\IEEEauthorblockA{\textit{dept. name of organization (of Aff.)} \\
%\textit{name of organization (of Aff.)}\\
%City, Country \\
%email address or ORCID}
%\and
%\IEEEauthorblockN{5\textsuperscript{th} Given Name Surname}
%\IEEEauthorblockA{\textit{dept. name of organization (of Aff.)} \\
%\textit{name of organization (of Aff.)}\\
%City, Country \\
%email address or ORCID}
%\and
%\IEEEauthorblockN{6\textsuperscript{th} Given Name Surname}
%\IEEEauthorblockA{\textit{dept. name of organization (of Aff.)} \\
%\textit{name of organization (of Aff.)}\\
%City, Country \\
%email address or ORCID}
%}
%%%%%%%%%%%%%%%%%%%%%%%%%%%%%%
%%%%%%%%%%%%%%%%%%%%%%%%%%%%%%%%%%%%%
\thispagestyle{empty}

\begin{huge}
	IEEE Copyright Notice
\end{huge}

\vspace{5mm} %5mm vertical space

\begin{large}
	Copyright (c) 2022 IEEE
\end{large}

\vspace{5mm} %5mm vertical space

\begin{large}
	Personal use of this material is permitted. Permission from IEEE must be obtained for all other uses, in any current or future media, including reprinting/republishing this material for advertising or promotional purposes, creating new collective works, for resale or redistribution to servers or lists, or reuse of any copyrighted component of this work in other works.
\end{large}

\vspace{5mm} %5mm vertical space

\begin{large}
	\textbf{Accepted to be published in:} IEEE ISCAS-2022; May 28 - June 1 , 2022.
	https://www.iscas2022.org/
	
\end{large}

\vspace{5mm} %5mm vertical space

%%%%%%%%%%%%%%%%%%%%%%%%%%

\maketitle

\begin{abstract}
Long short-term memory (LSTM) is a robust recurrent neural network architecture for learning spatiotemporal sequential data. However, it requires significant computational power for learning and implementing from both software and hardware aspects. This paper proposes a novel LiteLSTM architecture based on reducing the computation components of the LSTM using the weights sharing concept to reduce the overall architecture cost and maintain the architecture performance. The proposed LiteLSTM can be significant for learning big data where time-consumption is crucial such as the security of IoT devices and medical data. Moreover, it helps to reduce the CO2 footprint. The proposed model was evaluated and tested empirically on two different datasets from computer vision and cybersecurity domains. 
\end{abstract}

\begin{IEEEkeywords}
LiteLSTM, LSTM, GRU, MNIST, CO2 footprint
\end{IEEEkeywords}

\section{Introduction}

Sequential data modeling such as text, univariate time series, multivariate time series, audio signals, videos, genetic and amino acid sequences, and biological signals requires an apparatus that can recognize the temporal dependencies and relationships within the sequential data.
The recurrent neural network (RNN) was first designed in the early 1980s as the first neural network approach that targeted sequential data problems~\cite{deepLearnigBook}. 
The RNN has the capability to capture temporal dependencies due to its recurrent architecture in the sense that it recursively integrates the current new input into its self-previous output~\cite{graves2009novel}. Since it has an unrestricted but fading memory for the past, it can utilize the temporal dependencies to influence the learning of the structure within the data sequences~\cite{elsayed2019gateddissertation}.
The RNN has been applied in different research areas such as handwriting recognition~\cite{graves2009novel}, speech recognition~\cite{sak2014long,Graves2013SpeechRW}, language modeling~\cite{mikolov2010recurrent,mikolov2011extensions}, machine translation~\cite{bahdanau2014neural}, action recognition~\cite{du2015hierarchical}, stock prediction~\cite{kamijo1990stock}, video classification~\cite{yang2017tensor}, time series prediction~\cite{han2004prediction},and mental disorder prediction~\cite{petrosian2001recurrent}. 

However, the RNN has a significant weakness: its ability to learn long-term dependencies is limited due to the vanishing/exploding gradient problem. There are several attempts to solve the RNN major design problem and enhance its overall performance. The network loses the ability to learn when the error gradient is corrupted. To solve the vanishing/exploding gradient, extensions to the RNN architecture require adding an internal state (memory) that enforces a constant error flow through the RNN architecture stage. This constant error flow enhances the robustness of the error gradient over longer time scales. In addition, a gated control over the content of this internal state (memory) is also needed~\cite{hochreiter1997a}.

Nevertheless, this early LSTM model had major weaknesses. When it was first designed by Hochreiter and Schmidhuber~\cite{hochreiter1997a}, the LSTM model input data was assumed to be prior segmented into subsequences with explicitly marked ends that the memory could reset between each irreverent subsequences processing~\cite{hochreiter1997a,gers2000learning}. Moreover, this LSTM architecture did not have an internal reset component in case of processing continual input streams. Therefore, when the LSTM processes continuous input streams, the state action may grow infinitely and ultimately cause the LSTM architecture to fail~\cite{gers2000learning}.

In 2000,~\cite{gers2000learning} proposed a solution for the original LSTM problem that proposed in~\cite{hochreiter1997a}.~\cite{gers2000learning} added a forget gate beside the input and output gates into the LSTM architecture that resets the LSTM memory when the input is diversely different from the memory content and helps to remove the unnecessary information that the LSTM memory carries through the time. This LSTM approach~\cite{gers2000learning} (knows as standard LSTM or vanilla LSTM) is widely used to solve various problems such as speech recognition~\cite{sak2014long,soltau2016neural,chorowski2014end,miao2015eesen,graves2013hybrid}, language modeling~\cite{sundermeyer2012lstm,merity2017regularizing,sutskever2014sequence,miyamoto2016gated}, machine translation~\cite{cho2014properties,bahdanau2014neural,luong2014addressing,luong2015stanford}, time series classification~\cite{karim2018lstm,karim2018multivariate}, image segmentation~\cite{stollenga2015parallel,chen2018deeplab,reiter2006combined}, and video prediction~\cite{cho2014properties}.

However, this model also has critical weaknesses. First, the architecture does not have a direct connection from the memory state to the forget, input, and output gates. Hence, there is no control from the memory to the gates that could assist in preventing the gradient from vanishing or exploding. 
Second, the CEC does not have influential conduct over the forget and input gates when the output gate is closed, which could negatively affect the model due to the lack of primary information flow within the model~\cite{gers2002learning}.

To handle these problems in the standard LSTM, in 2002,~\cite{gers2002learning} added the peephole connections from the memory state cell to each of the LSTM forget, input, and output gates.
The peephole connections allowed the memory state to exert some control over the gates, which reinforces the LSTM architecture and prevents the lack of information flow through the model during the situation that leads to the output gate being closed~\cite{gers2002learning}.

The peephole added a generalization element to the standard LSTM. However, the major weakness of this architecture is that it becomes cost expensive due to the significant increase in the number of trainable parameters, memory, processing, and storage requirements to train the model and save the trained weights of the model and training time.

Nevertheless, the is a still-growing interest to study and apply the LSTM architecture to solve various sequential problems in different research domains due to the LSTM outperforms the GRU when in several tasks, when problems have large training datasets~\cite{greff2017lstm}. Moreover, Greff et al.~\cite{greff2017lstm} in 2017 showed that the LSTM exceeds the GRU performance in language modeling-related tasks. On the other hand, in some problems where the training datasets are small, the GRU outperforms the LSTM using a smaller computation budget~\cite{chung2014empirical}.

The era of big data requires robust tools to process large datasets. In addition, it requires accelerated time-consuming tools to process the data. Moreover, as the world tries to reduce the Carbon (CO2) footprint~\cite{bocken2012strategies} by reducing the usage of high-performance hardware, the LSTM implementation requirements cost is considered as one of the major LSTM drawbacks.

Spatiotemporal prediction problems are challenging to solve, utilizing only a gated recurrent architecture. Implementing such models is quite expensive from both resources and value aspects as a large number of parameters, rapid processors, large processing memory, and memory storage are needed. In addition, such models demand considerable time to train, validate and test. Moreover, implementing such a model for real-time training is a challenge. 

This paper attempts to evolve several computational aspects into a sophisticated performance level. This paper proposes a novel recurrent gated architecture: Lite long short-term memory (LiteLSTM) using one gate. The proposed LiteLSTM employes the concept of sharing weight among the gates introduced in the GRU~\cite{chung2014empirical} to reduce the model computation budget~\cite{elsayed2020reduced}. Also, it employs memory control over the gate using the peephole connection over the one gate. Besides, compared to the LSTM and GRU, the LiteLSTM has a smaller computation budget and implementation requirements, maintaining comparable accuracy. Furthermore, the LiteLSTM has a significant training time reduction compared to the LSTM due to the significant reduction in the computation budget.

%%%%%%%%%%%%%%%%%%%%%%%%%

\section{LiteLSTM Architecture}\label{LiteLSTM}

The proposed LiteLSTM aims to: reduce the overall implementation cost of the LSTM, solve the LSTM major problems, and maintain comparable accuracy performance to the LSTM. The architecture of the LiteLSTM consists of one trainable gated unit. We named the trainable gate the forget gate or network gate. The LiteLSTM has a peephole connection from the memory state to the forget gate, which preserves the memory state from the LSTM and keeps the CEC to avoid vanishing and/or exploding gradients.

\begin{figure}
	\centering
	\includegraphics[width=8cm,height=5cm]{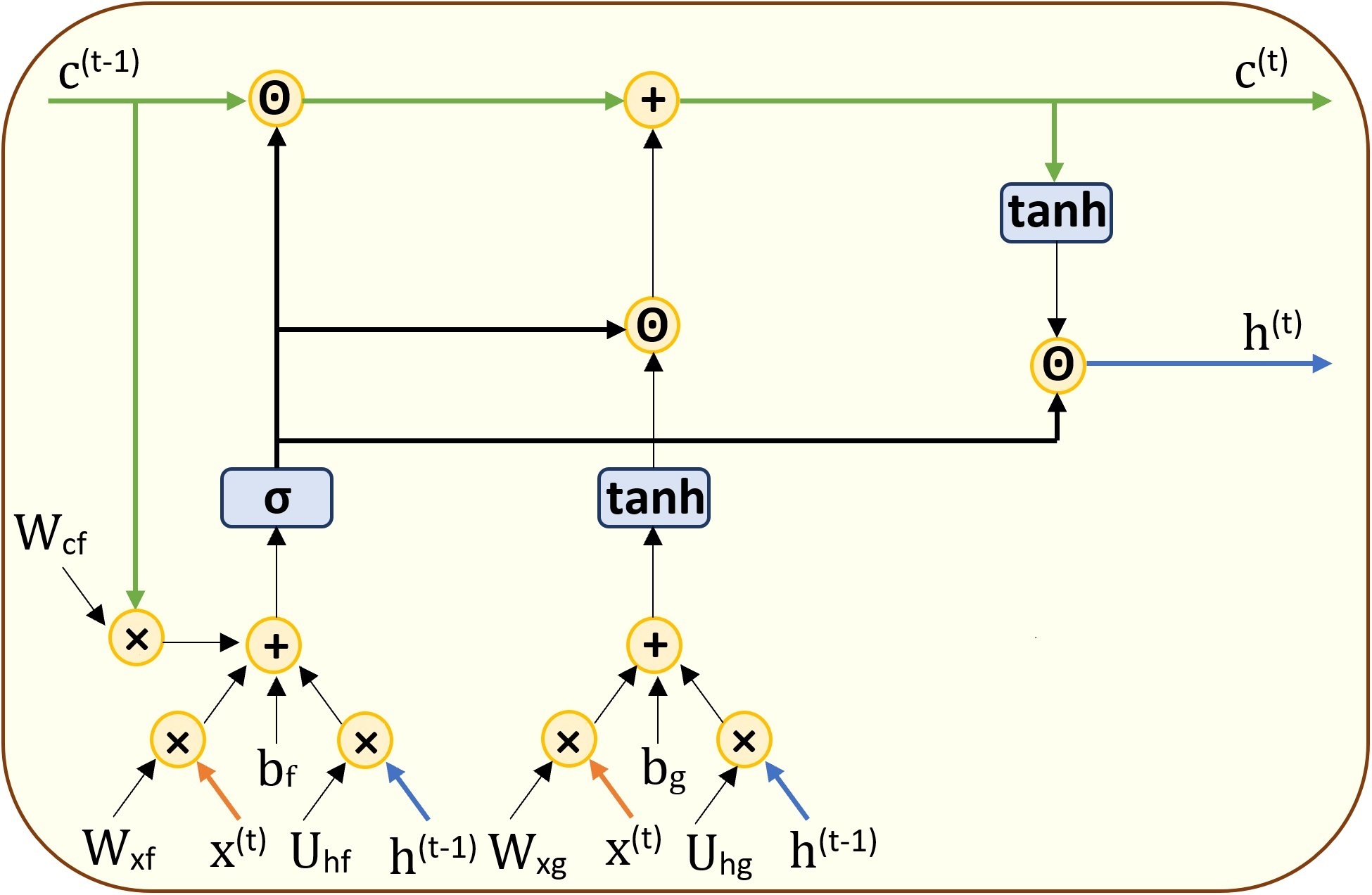}
	\caption{The operation level of the LiteLSTM architecture showing the weights of each component.}
	%	\caption{The proposed LiteLSTM block unrolled architecture.}
	\label{LiteLSTM_weights}
\end{figure}

Thus, the proposed LiteLSTM preserves the critical components of the LSTM as stated by~\cite{greff2017lstm} while reducing much of the parameter redundancy in the LSTM architecture. 
The LiteLSTM has a significant reduction in the number of trainable parameters that are required to implement the model. Therefore, the LiteLSTM reduced the training time, memory, and hardware requirements compared to the standard LSTM, peephole-based LSTM (pLSTM), and GRU architectures. 
Furthermore, the proposed LiteLSTM architecture preserves comparable prediction accuracy results to the LSTM. Figure~\ref{LiteLSTM_weights} shows a detailed architecture of the unrolled (unfolded) LiteLSTM assuming non-stacked input.

\begin{table*}
	\setlength{\tabcolsep}{2.5pt}
	\caption{Computational components comparison between the proposed LiteLSTM and the state-of-the-art recurrent architectures} % title of Table
	\centering % used for centering table
	\begin{tabular}{ l  l l  l l l} % centered columns (4 columns)
		\textbf{Comparison} &\textbf{RNN}&\textbf{GRU} & \textbf{LSTM} &\textbf{pLSTM}&\textbf{LiteLSTM}\\
		\hline	\\
		number of gates	& 0&	2	&	3& 3 & 1 \\
		number of activations  &1&	1	&	2 & 2 & 2\\
		state memory cell & $\times$	&	$\times$	&	 $\checkmark$	&  $\checkmark$ & $\checkmark$\\
		peephole connection& $\times$	& $\times$& $\times$& $\checkmark$& $\checkmark$ \\
		number of weight matrices& 2& 6 & 8 &11 & 6\\
		number of elementwise multiplication&2& 3 & 3 & 6 & 3\\
		number of bias vectors&1& 3  &  4 & 4 & 2\\
	\end{tabular}
	\label{comparison_tableable} % is used to refer this table in the text
\end{table*}
%The block contains the trainable network gate. 
%The model preserves the memory cell of the vanilla LSTM to process long data sequences and 
%keeps the CEC to manage the vanishing/exploding gradient problem. 
%We preserve the peephole connection between the memory cell and the forget gate but implement
%it with the convolution operation instead of the elementwise multiply. 

The LiteLSTM formulas are created as follows: During the forward pass within the LiteLSTM at time step $t$ the total input (inp), 
$inp^{(t)}$, to the 
single forget gate $f^{(t)}$ is calculated by

\begin{equation}\label{main_eqn}
	inp^{(t)} = \left[ W_{fx}, U_{fh}, W_{fc} \right] \left[x^{(t)}, h^{(t-1)}, c^{(t-1)}\right] + b_f
\end{equation}

\noindent
where $inp^{(t)}\in\mathbb{R}^{\eta\times 1}$, and $\eta\times 1$ is the of input vector $inp^{(t)}$.
$x^{(t)}$ is the input at time $t$, $ x^{(t)} \in \mathbb{R} ^{\eta  \times 1}$, $h^{(t-1)}$ is the output of the LiteLSTM architecture at time $t-1$, and the memory state cell at time $t-1$ denoted by $c^{(t-1)}$. Both $h^{(t-1)}, c^{(t-1)} \in \mathbb{R} ^{\eta  \times 1}$. $W_{fx}$, $U_{fh}$, and $W_{fc}$ are the weight sets. 
%$W_{fx}$ is $ \in \mathbb{R} ^{m \times r \times \gamma}$ and both %$U_{fh}$, and $W_{fc}$ are $ \in \mathbb{R} ^{m \times r \times %\kappa}$.
%$m$ and $r$ are the weights width and hight.
%For a given block, all kernels are the same size ($m\times m$)\@.
All three weight sets $W_{fx}$, $U_{fh}$, and $W_{fc}$ and biases $b_f$ are trainable. The square brackets indicate stacking. We will let $W_f = \left[ W_{fx}, U_{fh}, W_{fc} \right]$. %\in %\mathbb{R}^{m\times m \times \left(\gamma+2\kappa\right) \times n}$.
%where $n$ is the number of output channels.
In addition, we let $I_f = \left[x^{(t)}, h^{(t-1)}, c^{(t-1)}\right]$.% \in %\mathbb{R}^{\eta  \times \upsilon \times\left(\gamma+2\kappa\right)}$.
%The $W_f \times I_f \in \mathbb{R}^{\eta\times\upsilon\times n}$, and %$b_f \in \mathbb{R}^{n\times1}$.
%Note that the convolution operation between $W_{fc}$ and $c^{(t-1)}$ %represents a departure from
%Shi et al.~\cite{Shi2015} where an elementwise multiply was used (in %contrast to convolution).

%The total number of trainable parameters for the network gate is:

%\begin{equation}
	%	\label{eqnFgateParamCount}
	%	f_\mathrm{gate}^\# = \left(m^2(\gamma+2\kappa)+1\right)\cdot n.
	%\end{equation}
%\textbf{What is n? Stack height? Is it the same for all three weight sets?}
%\textbf{Identify all weights that are trainable}.

%The network gate value, $f_{gate}^{(t)} \in %\mathbb{R}^{\eta\times\upsilon\times n}$, 
%is obtained 
We apply a squashing function $G$ to the net input as follows:
\begin{equation}
	f_\mathrm{gate}^{(t)}= G(inp^{(t)}) .
\end{equation}
Depending on the application, the squashing function $G$ can be either the logistic sigmoid ($\sigma$) or hard sigmoid ($\mathrm{hardSig}$)~\cite{Gulcehre2016}. 
The values of $f^t$
%The output of $G$ 
falls in the range $(0,1)$ or $\left[0, 1\right]$, 
depending on using the logistic sigmoid ($\sigma$) or hard sigmoid function, respectively.			%Selecting either the logistic sigmoid ($\sigma$) or the hard sigmoid ($hardSig$) depends on the application.
Assuming that case of selecction the function as $\sigma$, the gate value $f^{t}$ is calculated by:

\begin{equation}
	%f^{(t)} = \sigma(W_{f}*[x^{(t)}, h^{(t-1)}, c^{(t-1)}] + b_{f}) .
	f^{(t)} = \sigma(W_{f}I_f + b_{f}) .
	\label{forget_gate_eqn}
\end{equation}
%where $x^{t}$ is the input to the LiteLSTM block at time $t$ and $h^{(t-1)}$ is the output image at time $t-1$. 
%$c^{(t)}$ is the memory cell of the block at time $t$. $b_{f}$ is the bias of the network gate. 
%$W_{f}$ is a stack of feedforward and recurrent propagation weights. 
%%The input and weights of the LiteLSTM are stacked. 
%\noindent
%Stacking makes the learning process more powerful than the non-stacked weights due to the influence of the $x^{t}$, $h^{(t-1)}$ and $c^{(t)}$ 
%across convolutional weight set. %It is empirically shown by 
%Lotter et al.~\cite{Lotter2017} and Heck et al.~\cite{heck2017simplified} show empirically
%that stacking the input for recurrent units achieves better results than non-stacked input. 

The input update (memory activation) equation is calculated by:
\begin{equation}
	\label{memoryActEqn}
	g^{(t)} =\tanh\left(W_g I_g + b_g\right)
\end{equation}

\noindent
where$W_g = \left[ W_{gx}, U_{gh} \right]$, %\in \mathbb{R}^{m  \times m %\times\left(\gamma+\kappa\right) \times n}$
and $I_g = \left[x^{(t)}, h^{(t-1)}\right]$% \in \mathbb{R}^{\eta  \times \upsilon \times\left(\gamma+\kappa\right)}$.
The dimension in $W_g$ is matching the dimension of the $W_f$ that maintains the dimension compaatability within the architecture design. %This approach is taken so that the dimension of $g_\mathrm{gate}^{(t)} %\in \mathbb{R}^{\eta\times\upsilon\times n}$
%matches the dimension of $f_\mathrm{gate}^{(t)}$.
%Similarly, the dimension of $b_g \in \mathbb{R}^{n\times 1}$ matches %that of $b_f \in \mathbb{R}^{n\times 1}$.
%Finally, the number of trainable parameters for the input update is

%\begin{equation}
	%	\label{eqnInputUpdateParamCount}
	%	g_\mathrm{update}^\# = \left(m^2(\gamma+\kappa)+1\right)\cdot n.
	%\end{equation}

%\noindent
%Eqns.~\ref{eqnFgateParamCount} and \ref{eqnInputUpdateParamCount} count the total number of trainable parameters for the LiteLSTM module/block, so
%the final count is given by

%\begin{equation}
	%	\label{parameters_count}
	%	\mathrm{LiteLSTM}_\mathrm{block}^\# = f_\mathrm{gate}^\# + g_\mathrm{update}^\#
	%\end{equation}

%The peephole connection of the memory state cell helps convey long interval information accumulated by 
%the previous LiteLSTM cells to the current cell so that we avoid vanishing/exploding gradients.
%The memory cell state, $c^{(t)}$, and output state, $h^{(t)}$, of the LSTM block are calculated as
%follows.
Finally, the LiteLSTM output is calculated by:

\begin{align}
	c^{(t)} &= f^{(t)} \odot c^{(t-1)} + f^{(t)} \odot g^{(t)}\label{memory_cell}\\
	h^{(t)} &= f^{(t)} \odot tanh(c^{(t)})\label{LiteLSTM_output_eqn}
\end{align}

\noindent
%The $\odot$ symbol denotes
%elementwise multiplication. 
%$\kappa$ is constrained to equal $n$ so that the dimensions match for the elementwise multiplication operations.

Table~\ref{comparison_tableable} shows a comparision between the architecure design and computation components of the RNN, standard LSTM, peephole-based LSTM (pLSTM), and the proposed LiteLSTM.

\section{Emperical Evaluatuation and Analysis}\label{emperical}
In this paper, the LiteLSTM has been empirically tested and evaluated in two different research domains: computer vision and anomaly detection in IoT. The MNIST~\cite{lecun1998mnist} is used as the computer vision experiment dataset, 
%the IMDB dataset~\cite{maas-EtAl:2011:ACL-HLT2011} used for the natural language processing task, 
and the IEEE IoT Network Intrusion Dataset~\cite{q70p-q449-19} is used for anomaly detection in IoT tasks. To perform our experiments, we used an Intel(R) Core(YM) i7-9700 CPU @3.00GHZ, 3000 Mhz processor, Microsoft Windows 10 OS, and 32 GB memory computer machine. We used Python 3.7.6, Keras 2.0.4, and Tensorflow 1.15.0.

\begin{figure*}%[b]
	\centering
	\includegraphics[width=16.5cm,height=3.5cm]{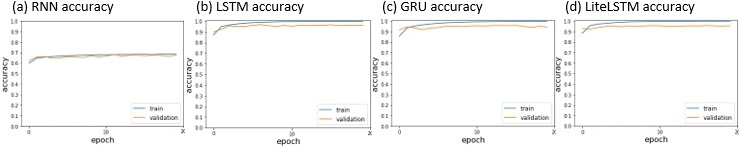}
	\caption{The accuracy diagrams of the recurrent architectures and LiteLSTM using MNIST dataset.}
	\label{mnist_accuracy_plots}
\end{figure*}

The first empirical evaluation of the LiteLSTM was performed using the MNIST dataset, which consists of $70,000$ images of handwritten digits between 0 and 9. The dataset is split into $60,000$ data samples for training and $10,000$ data samples for testing. The MNIST images were centered in a 28$\times$28 image by computing the center of mass of the pixels. 
%The model set 64-two layered architecture followed by a Softmax layer. 
The input was reshaped to map the LSTM framework. For the training process, the batch size was set to 128 and the number of epochs to 20. The Adam optimizer with learning rate $10^{-3}$, $\beta_{1} = 0.9$, $\beta_{2}= 0.999$, and $\epsilon=1e-07$.  Figure~\ref{mnist_accuracy_plots} shows the accuracy plots for each of the LiteLSTM and the state-of-the-art recurrent models. Table~\ref{mnist_result_Table} shows the accuracy results of the different recurrent architectures and the LiteLSTM, where the time is measured in minutes. The RNN shows a significantly shorter training time. However, it has the lowest performance compared to the other recurrent architectures. The LiteLSTM shows an improvement in accuracy compared to the other recurrent architectures.

\begin{table}
	\setlength{\tabcolsep}{2.5pt}
	\caption{Accuracy comparision between the LiteLSTM and the state-of-the-art recurrent architectures using MNIST dataset} % title of Table
	\centering % used for centering table
	\begin{tabular}{ l  l  ll ll} % centered columns (4 columns)
		\textbf{Comparision} & \textbf{RNN}& \textbf{GRU}&\textbf{LSTM}&\textbf{pLSTM}&\textbf{LiteLSTM} \\%\textbf{GRU} &
		\hline	\\
		\textbf{Time(m)}& \textbf{11.24}& \textbf{43.01}& 60.36& 75.45& \textbf{42.94}\\%54.51 &
		\textbf{Parameters}& \textbf{792,210}&812,610&822,810& 833,010& 812,610 \\ %812,610 &
		\textbf{Accuracy(\%)}& 67.64& 94.09 &95.70&95.99&\textbf{96.07}\\%94.09&
		
	\end{tabular}
	\label{mnist_result_Table} % is used to refer this table in the text
\end{table}

The second empirical evaluation of the LiteLSTM was performed using the IEEE IoT Network Intrusion Dataset. The dataset consists of 42 raw network packet files (pcap) at different time points. The IoT devices, namely SKT NUGU (NU 100) and EZVIZ Wi-Fi camera (C2C Mini O Plus 1080P), were used to generate traffic for IoT devices. The data contains normal traffic flow and different types of cyberattacks, namely: ARP spoofing attack, DoS (SYN flooding) attack, scan (host and port scan) attack, scan(port and OS scan) attack, (UDP/ACK/HTTP Flooding) of zombie PC compromised by Mirai malware, Mirai-ACK flooding attack, Mirai-HTTP flooding attack, and Telnet brute-force attack. In our experiments, we used a dataset to experiment with the LiteLSTM twice: first to detect whether an attack occurred or not (as binary dataset) and another experiment to detect the type of the attack. We set the batch size to 32 and the number of epochs to 20. Table~\ref{IoT_results_binary_table} shows the binary experimental results for the LiteLSTM and the recurrent architectures. Table~\ref{IoTmulti_results} shows the detection results of the LiteLSTM and the recurrent architectures for detecting different types of cyberattacks.

\begin{table}
	\setlength{\tabcolsep}{2.5pt}
	\caption{Accuracy comparision between the LiteLSTM and the state-of-the-art recurrent architectures using IEEE IoT Network Intrusion Binary Dataset dataset} % title of Table
	\centering % used for centering table
	\begin{tabular}{ l  l  ll  l l} % centered columns (4 columns)
		\textbf{Comparison} &\textbf{RNN}&\textbf{GRU}&\textbf{LSTM}&\textbf{pLSTM}&\textbf{LiteLSTM} \\%\textbf{GRU}
		\hline	\\
		\textbf{Time (m)} & \textbf{2.02}&4.26 & 5.42& 6.78& 4.38 \\%4.38
		\textbf{Precision}& 81.44&93.28 & 94.22 & \textbf{96.53} & 93.82 \\ %93.28
		\textbf{Recall}& 97.63&97.57& 94.84 & 95.45 &\textbf{98.34}\\%97.57
		\textbf{F1-score} & 88.80&91.34 & 95.97& 95.99&\textbf{96.03}\\%91.34
		\textbf{Accuracy(\%)}& 98.7&99.51 & 99.50 & 99.56 & \textbf{99.60}\\%99.51
		
	\end{tabular}
	\label{IoT_results_binary_table} % is used to refer this table in the text
\end{table}

\begin{table}
	\setlength{\tabcolsep}{2.5pt}
	\caption{Accuracy comparision between the LiteLSTM and the state-of-the-art recurrent architectures using IEEE IoT network intrusion detection for different cyberattacks dataset} % title of Table
	\centering % used for centering table
	\begin{tabular}{ l  l  ll  l l} % centered columns (4 columns)
		\textbf{Comparison} &\textbf{RNN}&\textbf{GRU}&\textbf{LSTM}&\textbf{pLSTM}&\textbf{LiteLSTM} \\%\textbf{GRU}
		\hline	\\
		\textbf{Time (m) }&  \textbf{2.033} & 5.15& 5.52 & 6.90  &  5.15 \\%4.47 
		\textbf{F1-score}       & 88.89&91.34  & 91.19  & 91.38 &  \textbf{91.53 }\\% 91.34
		\textbf{Accuracy(\%)}       &  83.35 &  86.70& 86.90 & 87.03 &  \textbf{87.10} \\% 86.70
		
	\end{tabular}
	\label{IoTmulti_results} % is used to refer this table in the text
\end{table}
\section{Conclusion}
The proposed LiteLSTM architecture novelty lies in the following aspects. First, the LiteLSTM consists of one gate that serves as a multifunctional gate. Thus, the overall number of training parameters is reduced approximately by one-third of the LSTM or the peephole-LSTM. In addition, maintaining the peephole connection from the memory state cell to the existing gate maintains the control of the memory over the gate in contrast to the LSTM. The overall budget of implementing the LiteLSTM, including the training time, memory footprint, memory storage, and processing power, is smaller than the LSTM. We empirically evaluated the LiteLSTM using two datasets: MNIST and IEEE IoT Network Intrusion Detection datasets. Due to the optimized LiteLSTM architecture design, we were able to complete the empirical tasks using a computer processor without involving the GPU in the computational process. Thus, the LiteLSTM architecture helps to reduce the CO2 footprint. The LiteLSTM outperforms the accuracy of the recurrent architecture in different application domains. The proposed LiteLSTM architecture is an attractive candidate for future hardware implementation on small and portable devices, especially IoT devices.

\bibliographystyle{ieeetr}
\bibliography{referencesLiteLSTM}

\begin{thebibliography}{10}

\bibitem{deepLearnigBook}
I.~Goodfellow, Y.~Bengio, and A.~Courville, {\em Deep Learning}.
\newblock MIT Press, 2016.
\newblock \url{http://www.deeplearningbook.org}.

\bibitem{graves2009novel}
A.~Graves, M.~Liwicki, S.~Fern{\'a}ndez, R.~Bertolami, H.~Bunke, and
  J.~Schmidhuber, ``A novel connectionist system for unconstrained handwriting
  recognition,'' {\em IEEE Transactions on Pattern Analysis and Machine
  Intelligence}, vol.~31, no.~5, pp.~855--868, 2009.

\bibitem{elsayed2019gateddissertation}
N.~Elsayed, {\em Gated convolutional recurrent neural networks for predictive
  coding}.
\newblock University of Louisiana at Lafayette, 2019.

\bibitem{sak2014long}
H.~Sak, A.~Senior, and F.~Beaufays, ``Long short-term memory recurrent neural
  network architectures for large scale acoustic modeling,'' in {\em Fifteenth
  annual conference of the international speech communication association},
  2014.

\bibitem{Graves2013SpeechRW}
A.~Graves, A.~rahman Mohamed, and G.~E. Hinton, ``Speech recognition with deep
  recurrent neural networks,'' {\em 2013 IEEE International Conference on
  Acoustics, Speech and Signal Processing}, pp.~6645--6649, 2013.

\bibitem{mikolov2010recurrent}
T.~Mikolov, M.~Karafi{\'a}t, L.~Burget, J.~{\v{C}}ernock{\`y}, and
  S.~Khudanpur, ``Recurrent neural network based language model,'' in {\em
  Eleventh Annual Conference of the International Speech Communication
  Association}, 2010.

\bibitem{mikolov2011extensions}
T.~Mikolov, S.~Kombrink, L.~Burget, J.~{\v{C}}ernock{\`y}, and S.~Khudanpur,
  ``Extensions of recurrent neural network language model,'' in {\em Acoustics,
  Speech and Signal Processing (ICASSP), 2011 IEEE International Conference
  on}, pp.~5528--5531, IEEE, 2011.

\bibitem{bahdanau2014neural}
D.~Bahdanau, K.~Cho, and Y.~Bengio, ``Neural machine translation by jointly
  learning to align and translate,'' {\em arXiv preprint arXiv:1409.0473},
  2014.

\bibitem{du2015hierarchical}
Y.~Du, W.~Wang, and L.~Wang, ``Hierarchical recurrent neural network for
  skeleton based action recognition,'' in {\em Proceedings of the IEEE
  conference on computer vision and pattern recognition}, pp.~1110--1118, 2015.

\bibitem{kamijo1990stock}
K.-i. Kamijo and T.~Tanigawa, ``Stock price pattern recognition-a recurrent
  neural network approach,'' in {\em Neural Networks, 1990., 1990 IJCNN
  International Joint Conference on}, pp.~215--221, IEEE, 1990.

\bibitem{yang2017tensor}
Y.~Yang, D.~Krompass, and V.~Tresp, ``Tensor-train recurrent neural networks
  for video classification,'' in {\em International Conference on Machine
  Learning}, pp.~3891--3900, PMLR, 2017.

\bibitem{han2004prediction}
M.~Han, J.~Xi, S.~Xu, and F.-L. Yin, ``Prediction of chaotic time series based
  on the recurrent predictor neural network,'' {\em IEEE Transactions on Signal
  Processing}, vol.~52, no.~12, pp.~3409--3416, 2004.

\bibitem{petrosian2001recurrent}
A.~Petrosian, D.~Prokhorov, W.~Lajara-Nanson, and R.~Schiffer, ``Recurrent
  neural network-based approach for early recognition of alzheimer's disease in
  {EEG},'' {\em Clinical Neurophysiology}, vol.~112, no.~8, pp.~1378--1387,
  2001.

\bibitem{hochreiter1997a}
S.~Hochreiter and J.~Schmidhuber, ``Long short-term memory,'' {\em Neural
  Computation}, vol.~9, no.~8, pp.~1735--1780, 1997.

\bibitem{gers2000learning}
F.~A. Gers, J.~Schmidhuber, and F.~Cummins, ``Learning to forget: Continual
  prediction with {LSTM},'' {\em Neural Computation}, pp.~2451--2471, 2000.

\bibitem{soltau2016neural}
H.~Soltau, H.~Liao, and H.~Sak, ``Neural speech recognizer: Acoustic-to-word
  {LSTM} model for large vocabulary speech recognition,'' {\em arXiv preprint
  arXiv:1610.09975}, 2016.

\bibitem{chorowski2014end}
J.~Chorowski, D.~Bahdanau, K.~Cho, and Y.~Bengio, ``End-to-end continuous
  speech recognition using attention-based recurrent {NN}: first results,''
  {\em arXiv preprint arXiv:1412.1602}, 2014.

\bibitem{miao2015eesen}
Y.~Miao, M.~Gowayyed, and F.~Metze, ``{EESEN}: End-to-end speech recognition
  using deep {RNN} models and {WFST}-based decoding,'' in {\em Automatic Speech
  Recognition and Understanding (ASRU), 2015 IEEE Workshop on}, pp.~167--174,
  IEEE, 2015.

\bibitem{graves2013hybrid}
A.~Graves, N.~Jaitly, and A.-r. Mohamed, ``Hybrid speech recognition with deep
  bidirectional {LSTM},'' in {\em Automatic Speech Recognition and
  Understanding (ASRU), 2013 IEEE Workshop on}, pp.~273--278, IEEE, 2013.

\bibitem{sundermeyer2012lstm}
M.~Sundermeyer, R.~Schl{\"u}ter, and H.~Ney, ``{LSTM} neural networks for
  language modeling,'' in {\em Thirteenth annual conference of the
  international speech communication association}, 2012.

\bibitem{merity2017regularizing}
S.~Merity, N.~S. Keskar, and R.~Socher, ``Regularizing and optimizing {LSTM}
  language models,'' {\em arXiv preprint arXiv:1708.02182}, 2017.

\bibitem{sutskever2014sequence}
I.~Sutskever, O.~Vinyals, and Q.~V. Le, ``Sequence to sequence learning with
  neural networks,'' in {\em Advances in neural information processing
  systems}, pp.~3104--3112, 2014.

\bibitem{miyamoto2016gated}
Y.~Miyamoto and K.~Cho, ``Gated word-character recurrent language model,'' {\em
  arXiv preprint arXiv:1606.01700}, 2016.

\bibitem{cho2014properties}
K.~Cho, B.~Van~Merri{\"e}nboer, D.~Bahdanau, and Y.~Bengio, ``On the properties
  of neural machine translation: Encoder-decoder approaches,'' {\em arXiv
  preprint arXiv:1409.1259}, 2014.

\bibitem{luong2014addressing}
M.-T. Luong, I.~Sutskever, Q.~V. Le, O.~Vinyals, and W.~Zaremba, ``Addressing
  the rare word problem in neural machine translation,'' {\em arXiv preprint
  arXiv:1410.8206}, 2014.

\bibitem{luong2015stanford}
M.-T. Luong and C.~D. Manning, ``Stanford neural machine translation systems
  for spoken language domains,'' in {\em Proceedings of the International
  Workshop on Spoken Language Translation}, pp.~76--79, 2015.

\bibitem{karim2018lstm}
F.~Karim, S.~Majumdar, H.~Darabi, and S.~Chen, ``{LSTM} fully convolutional
  networks for time series classification,'' {\em IEEE Access}, vol.~6,
  pp.~1662--1669, 2018.

\bibitem{karim2018multivariate}
F.~Karim, S.~Majumdar, H.~Darabi, and S.~Harford, ``Multivariate {LSTM-FCN}s
  for time series classification,'' {\em arXiv preprint arXiv:1801.04503},
  2018.

\bibitem{stollenga2015parallel}
M.~F. Stollenga, W.~Byeon, M.~Liwicki, and J.~Schmidhuber, ``Parallel
  multi-dimensional {LSTM}, with application to fast biomedical volumetric
  image segmentation,'' in {\em Advances in neural information processing
  systems}, pp.~2998--3006, 2015.

\bibitem{chen2018deeplab}
L.-C. Chen, G.~Papandreou, I.~Kokkinos, K.~Murphy, and A.~L. Yuille, ``Deeplab:
  Semantic image segmentation with deep convolutional nets, atrous convolution,
  and fully connected crfs,'' {\em IEEE Transactions on Pattern Analysis and
  Machine Intelligence}, vol.~40, no.~4, pp.~834--848, 2018.

\bibitem{reiter2006combined}
S.~Reiter, B.~Schuller, and G.~Rigoll, ``A combined {LSTM-RNN-HMM}-approach for
  meeting event segmentation and recognition,'' in {\em Acoustics, Speech and
  Signal Processing, 2006. ICASSP 2006 Proceedings. 2006 IEEE International
  Conference on}, vol.~2, pp.~II--II, IEEE, 2006.

\bibitem{gers2002learning}
F.~A. Gers, N.~N. Schraudolph, and J.~Schmidhuber, ``Learning precise timing
  with {LSTM} recurrent networks,'' {\em Journal of Machine Learning Research},
  vol.~3, pp.~115--143, 2002.

\bibitem{greff2017lstm}
K.~Greff, R.~K. Srivastava, J.~Koutn{\'\i}k, B.~R. Steunebrink, and
  J.~Schmidhuber, ``{LSTM}: A search space odyssey,'' {\em IEEE Transactions on
  Neural Networks and Learning Systems}, vol.~28, no.~10, pp.~2222--2232, 2017.

\bibitem{chung2014empirical}
J.~Chung, C.~Gulcehre, K.~Cho, and Y.~Bengio, ``Empirical evaluation of gated
  recurrent neural networks on sequence modeling,'' {\em arXiv preprint
  arXiv:1412.3555}, 2014.

\bibitem{bocken2012strategies}
N.~M. Bocken and J.~M. Allwood, ``Strategies to reduce the carbon footprint of
  consumer goods by influencing stakeholders,'' {\em Journal of Cleaner
  Production}, vol.~35, pp.~118--129, 2012.

\bibitem{elsayed2020reduced}
N.~Elsayed, A.~S. Maida, and M.~Bayoumi, ``Reduced-gate convolutional long
  short-term memory using predictive coding for spatiotemporal prediction,''
  {\em Computational Intelligence}, vol.~36, no.~3, pp.~910--939, 2020.

\bibitem{Gulcehre2016}
C.~Gulcehre, M.~Moczulski, M.~Denil, and Y.~Bengio, ``Noisy activation
  functions,'' in {\em International Conference on Machine Learning},
  pp.~3059--3068, 2016.

\bibitem{lecun1998mnist}
Y.~LeCun, ``The mnist database of handwritten digits,'' {\em http://yann.
  lecun. com/exdb/mnist/}, 1998.

\bibitem{q70p-q449-19}
H.~Kang, D.~H. Ahn, G.~M. Lee, J.~D. Yoo, K.~H. Park, and H.~K. Kim, ``Iot
  network intrusion dataset,'' 2019.

\end{thebibliography}

\end{document}